# A comprehensive overview of deep learning models for object detection from videos/images

## Sukana Zulfqar

Department of Computer Science,
University of Agriculture Faisalabad,
Faisalabad, 38000, Punjab, Pakistan
Email: sukainazulfqar44@gmail.com

## Sadia Saeed

Faculty of Information Technology and Computer Science (FoIT&CS),
University of Central Punjab,
Lahore, 54783, Punjab, Pakistan
Email: sadiasaeed124@gmail.com

## M. Azam Zia

Department of Computer Science,
University of Agriculture Faisalabad,
Faisalabad, 38000, Punjab, Pakistan
Email: mazamzia@uaf.edu.pk

## Anjum Ali

Department of Software Engineering,
Riphah International University,
Faisalabad Campus, Pakistan
Email: malik.anjumali@gmail.com

## Faisal Mehmood*

Department of Civil and Transportation Engineering,
Shenzhen University,
Shenzhen 518060, China
Email: faisalmehmood685@uaf.edu.pk
*Corresponding author

## Abid Ali

Department of Computer Science,
University of Agriculture Faisalabad,
Faisalabad, 38000, Punjab, Pakistan
Email: abidali457287@gmail.com

**Abstract:** Object detection in video and image surveillance is a well-established yet rapidly evolving task, strongly influenced by recent deep learning advancements. This review summarises modern techniques by examining architectural innovations, generative model integration, and the use of temporal information to enhance robustness and accuracy. Unlike earlier surveys, it classifies methods based on core architectures, data processing strategies, and surveillance specific challenges such as dynamic environments, occlusions, lighting variations, and real-time requirements. The primary goal is to evaluate the current effectiveness of semantic object detection, while secondary aims include analysing deep learning models and their practical applications. The review covers CNN-based detectors, GAN-assisted approaches, and temporal fusion methods, highlighting how generative models support tasks such as reconstructing missing frames, reducing occlusions, and normalising illumination. It also outlines preprocessing pipelines, feature extraction progress, benchmarking datasets, and comparative evaluations. Finally, emerging trends in low-latency, efficient, and spatiotemporal learning approaches are identified for future research.

**Biographical notes:** Sukana Zulfqar is a PhD scholar at the Harbin Institute of Technology (HIT), Shenzhen, focusing on machine learning and computer vision.

Sadia Saeed is pursuing her PhD at the University of Central Punjab, Lahore, with interests in AI and data science.

M. Azam Zia is a Professor at the University of Agriculture Faisalabad, specialising in data analytics and computational intelligence.

Anjum Ali is a Lecturer at Riphah International University, Faisalabad Campus.

Faisal Mehmood is a Postdoctoral Researcher at Shenzhen University working on deep learning and computer vision.

Abid Ali holds an MS from UAF and works in machine learning and pattern recognition.

---

# 1 Introduction

The objective of computer vision's object detection and recognition has been the localisation of objects within an image. In many applications, such as geographic resource mapping, crop harvest analysis, disaster management, traffic planning, and navigation, object detection is an essential task (Kaur and Singh, 2023; Zohar et al., 2023; Diwan et al., 2023). The environment we live in is filled with a variety of objects. It is a difficult effort to identify those objects with a machine's assistance. A subfield of computer science called computer vision gives machines the ability to recognise, classify, and interact with objects in still and moving pictures. A video is an uninterrupted series of pictures, or 'video frames', that are shown at a particular frame rate. Because of the interconnection of brain neurons, humans are able to instantly identify or recognise objects in pictures or videos and pinpoint their location. Video object detection is a task involving artificial intelligence that employs the identical method to identify objects. Object detection is an essential component of computer vision, and for several decades it has been a significant and ongoing area of research (Kaur and Singh, 2023; Wang et al., 2023). Finding and following moving images in different types of media is an important part of many uses, such as video security (Yadav et al., 2023), intelligent traffic management (Gong et al., 2023), and digital city infrastructure (Shepelev et al., 2020). These jobs can be done by closed-circuit televisions (CCTVs) (Mirzaei et al., 2023), aircraft (Gaffey et al., 2022), unmanned aerial vehicles (UAVs) (Zhou et al., 2018; Mahmood et al., 2019), and even moving nanoparticles through soil to protect the environment (de Vries et al., 2022).

The two parts of object detection are object detection in pictures and object detection in videos. Object recognition is an interesting area of computer vision that has been used in many situations, including finding people on the street (Murthy et al., 2022), faces (Chen et al., 2020a), self-driving cars (Chen et al., 2020b), and robotic vision (Li et al., 2023). Deep learning methods are always getting better, which means that more and more object detection tasks can be done successfully (Bengio et al., 2021).

Video object detection finds things using video data instead of static images, which is what traditional object detection does. Self-driving cars (Bateni et al., 2020) and video surveillance (Guillermo et al., 2020) are two major uses of video object recognition that have helped it grow. One of the new tasks for the ImageNet Large Scale Visual Recognition Challenge (ILSVRC2015) in 2015 was video object detection (Guillermo et al., 2020). There has been a surge in research into video object detection due to ILSVRC2015.

Previous attempts at object detection in video entailed identifying objects in each frame of the image. The two main types of object detection methods are one-stage detectors and two-stage detectors. In many cases, one-stage detectors (Guillermo et al., 2020) outperform two-stage detectors (Diwan et al., 2023) in terms of computing efficiency. Nevertheless, it has been demonstrated that two-stage detectors outperform one-stage detectors in terms of accuracy.

There are a number of characteristics present in video data that are ignored when object detection is applied to individual frames: redundancies in feature extraction across neighbouring frames occur as a result of:

1   Spatial and temporal correlations among picture frames. Computational inefficiency results from feature detection in each frame.

2   Occlusion, motion blur, defocus, and position changes might cause some frames in a protracted video stream to have low quality (Ding et al., 2023).

Low accuracies are the result of object detection from low quality frames. Methods for video object identification aim to overcome these obstacles. In order to increase precision, some methods leverage the spatial-temporal data, for as merging features at multiple levels (Chen et al., 2020c). Other methods aim to improve detection efficiency while decreasing information redundancy, for instance (Delibaşoğlu, 2023; Khalid et al., 2025).

Feature extraction from manually created videos has been around for a while (Majumder et al., 2020; Wang et al., 2020). Computer vision (Li et al., 2019; Fang et al., 2023; Mehmood et al., 2024a), speech processing (Tao and Busso, 2020; Wei and Kehtarnavaz, 2020), and multi-modality signal processing (Chen et al., 2017) are just a few areas where deep learning models have significantly outperformed more traditional methods, thanks to the proliferation of convolutional neural networks and other forms of deep learning. The ILSVRC2015 competition sparked the development of several video object identification methods based on deep learning. Online instruction is the standard. Even with sophisticated networks, testing on recent GPUs can achieve the 30 fps video rate (Poggi et al., 2022), paving the way for networks to be deployed in real-time.

## 2   Benefits of object detection

- Biometrics plays a crucial role in recognising individual identities and ensuring security, as technology advances. Biometric authentication provides a more dependable means of verifying an individual's identity. The authentication is conducted by utilising various biological characteristics of each individual, including fingerprint, DNA, retina, auditory, and so forth (Yuan and Lu, 2018). Previous studies have employed a variety of object detecting methods for biometric analysis.

- Technological video surveillance systems have been implemented in the majority of inhabited areas (metropolises, parks, schools, shopping centres, etc.) due to the impracticability of human beings to perpetually monitor video clips. Video surveillance relies heavily on object detection to simultaneously identify and monitor instances of a specific object in a scene (Dai et al., 2022; Elhoseny, 2020). For instance, object detection may be used to track a suspect vehicle or person from the video.

- Autonomous robotics have emerged as a highly intriguing field of research in recent times. The principal function of the robot is object detection, which enables it to identify adjacent objects and execute operations such as delivering information, opening and closing doors, sounding an alarm, and more (Chatterjee et al., 2020).

- Human detection presents a significant challenge in the field of computer vision due to the diverse appearances and extensive range of poses that people assume as objects. To find people in pictures or videos, different object detection architectures have been suggested, such as pedestrian recognition (Hasan et al., 2021). Object identification has allowed for the rapid counting of crowds in densely crowded places such as parks, malls, etc. (Kaur and Singh, 2022).

People have used object detection to find a specific face, which is the first use case for human object detection (Modwel et al., 2021). Face recognition is very good at finding things while taking very little time to compute. Object detection has been used in a lot of different ways thanks to face recognition. This idea is already being used in a lot of apps to find smiles in real time using cameras, face makeup, age calculations, and other things (Nguyen et al., 2019). Research into smart vehicle systems, such as the driverless car, has been difficult due to technological advancements (Wei et al., 2023). In order to manage the vehicle's speed and detect, locate, or follow surrounding objects, these smart-systems are necessary. Images with a finer grain and a regional level, such as those used for traffic light detection and recognition, are likewise compatible with the object detection system (Santra et al., 2019).

## 3 Datasets for object detection

A dataset is a group of appropriate information (text, image and video) compiled in a particular manner in accordance with the specifications of the algorithm. A logically and mathematically developed model is necessary for artificially trained machines. The training procedure aids in the development of these models. The two main elements of the training process are the technique and the dataset. The dataset aids in the machine's learning process from the input samples, and the approach specifies the learning phases the machine will take. The number and quality of the dataset have a significant impact on the recognition model's performance.

### 3.1 Image-based object detection datasets

#### 3.1.1 Pascal VOC

In computer vision technology, PASCAL VOC is a prevalent dataset that has been used to test and create new algorithms. Numerous tasks, including object detection, image classification, activity detection, and segmentation, are carried out by those algorithms (Everingham et al., 2006, 2010, 2015). Between 2005 and 2012, there were PASCAL VOC challenges. Each year, new versions were created, however two of the most popular versions are VOC07 and VOC12. Every picture was collected from Flickr or the current data source. There were just four classes identified by VOC 2005: humans, automobiles, bikes, and motorcycles. However, in order to identify 20 item classes from photos, the PASCAL VOC 12 dataset has 27,450 annotated objects and 11,530 training images. A dataset representing everyday objects encountered by humans was assembled through the selection and annotation of 20 object classes:

a   person: individual

b   indoor: chair, dining table, sofa, potted plant, television/monitor c   vehicle: bus, bicycle, car, boat, motorcycle, aeroplane, and train a   animal: cat, dog, cow, bird, sheep and horse.

**Table 1**   Image-based object detection dataset statistics

| Dataset | Years | Classes (common objects) | Annotated objects | Training images |
|---------|-------|--------------------------|-------------------|-----------------|
| VOC05 | 2005–2012 | Humans, automobiles, bikes, motorcycles | - | - |
| VOC12 | 2005–2012 | Person, indoors (sofa, TV/monitor, bottle, dining table, chair), transport (bus, bicycle, automobile, boat, motorcycle, aircraft, train), (horse, sheep, dog, cow, bird, cat) | 27,450 | 11,530 |

### 3.1.2  *ImageNet Large Scale Visual Recognition Challenge*

ImageNet Large Scale Visual Recognition Challenge (ILSVRC) is an annual computer vision challenge that took place from 2010 to 2017 (Deng et al., 2009). Publicly accessible datasets, challenges, and associated workshops were all part of the ILSVRC. For detecting problems, it gathered photos from ImageNet (via the WorldNet) hierarchy (Deng, 2009). It only began working on the 1,000-category picture classification problem in 2010. The advancement of technology and time has brought out new challenges. The dataset was created in 2017 with the goal of identifying 30 fully labelled groups that were distinguished by a number of parameters, including the average number of object instances, movement type, and degree of video clutter. There are more than 15 million high-resolution pictures in the image dataset (Deng et al., 2009; Faisal et al., 2025).

### 3.1.3  *MS COCO dataset*

There are 91 common object categories in the Microsoft Common Objects in Context Collection. More than 5,000 examples have been tagged in 82 of these categories. The collection has a total of 2,500,000 instances that have been marked up and 328,000 pictures. While ImageNet is a well-known dataset, COCO differs in that it has fewer categories but more instances within each category. This can help you learn detailed object models that can be localised precisely in two dimensions. Out of all the datasets, PASCAL VOC (Santra et al., 2019) and SUN (Deng et al., 2009) have the lowest number of occurrences per category compared to this one. Furthermore, one significant difference between our dataset and others is the quantity of labelled instances per image, which may aid in understanding contextual information.

**Table 2**    Image-based dataset explanation.

| Feature | Description |
|---|---|
| Dataset name | Microsoft Common Objects in Context (MS COCO) |
| Object categories | 91 common object categories |
| Instances per category | 82 categories have more than 5,000 labelled instances |
| Labelled instances | 2,500,000 instances in 328,000 images |
| Object instance distribution | More instances per category compared to ImageNet, PASCAL VOC, and SUN datasets |
| Learning benefits | Aids in learning detailed object models with precise 2D localisation |
| Comparison with ImageNet | Fewer categories but more instances per category, allowing for more detailed object model learning |
| Dataset size comparison | Significantly larger number of instances per category compared to PASCAL VOC and SUN datasets |
| Critical distinction | Number of labelled instances per image is a critical distinction, facilitating the learning of contextual information |

### 3.1.4  *OID dataset*

The Open Images Dataset (OID) V4 comprises 9.2 million images accompanied by standardised annotations for three tasks: visual relation detection, object detection, and image classification (Kuznetsova et al., 2020a). The photographs have been acquired from Flicker without utilising predetermined class names. The dataset comprises 30.1 million image-level annotations, of which 19.8 thousand are concepts, 375 visual relationship annotations for 57 classes, and 15.4 million bounding boxes for 600 object classes. Because expert annotators labelled each object, the dataset offers accuracy and consistency. This dataset's photos are all of excellent quality and include many objects.

**Table 3**     Open images (OID) V4 description

| Feature | Description |
| --- | --- |
| Dataset name | Open Images (OID) V4 |
| Tasks | Visual relation detection, object detection, image classification |
| Number of images | 9.2 million |
| Source | Acquired from Flicker without predetermined class names |
| Image-level annotations | 30.1 million annotations |
| Concepts | 19.8 thousand concepts |
| Classes (object detection) | 600 object classes |
| Bounding boxes | 15.4 million bounding boxes |
| Visual relationships | 375 annotations |
| Image quality | Excellent quality |
| Annotation accuracy | Expert annotators labelled each object, providing accuracy and consistency |

### 3.2   Video-based object detection datasets

A lot of information has been made available for different uses (Deng et al., 2009; Lin et al., 2014). ImageNet VID (Kuznetsova et al., 2020a) is a frequently employed dataset utilised for video object detection and frequently serves as a standard for video object recognition. The dataset is partitioned into two distinct sets: a training set comprising 3,862 video samples, and a validation set comprising 555 video samples. Each frame of the video broadcasts is annotated at a rate of either 25 or 30 frames per second. Moreover, out of the total number of object categories in the ImageNet DET dataset, 30 are present in this dataset (Rybak and Dudczyk, 2020).

### 3.2.1   ImageNet VID dataset

When contrasted with datasets that are utilised for the detection of static picture objects, such as COCO, the amount of objects that are present in each frame of the ImageNet VID dataset is somewhat lesser. Despite its popularity, the ImageNet VID dataset does not adequately capture the impact of all video object recognition techniques. A large-scale dataset called YouTube-Bounding Boxes (YT-BB) was made available in Real et al. (2017). It contains tight bounding boxes and high accuracy classification labels that were manually annotated at a rate of one frame per second on YouTube video samples. A portion of the COCO label set, YT-BB has over 380,000 video clips with 5.6 million bounding boxes of 23 different object types. Nevertheless, because the images were taken with hand-held mobile phones, the dataset only has 23 object categories and has relatively poor image quality.

**Table 4** Video-based object detection description

| Dataset | Description | Number of video clips | Number of bounding boxes | Object categories | Image quality |
|---------|-------------|----------------------|--------------------------|-------------------|---------------|
| YouTube-Bounding Boxes | Large-scale video dataset | 380,000 | 5.6 million | 23 | Relatively poor quality |

**Table 5** Video-based object detection dataset description

| Dataset name | Description | Frames | Bounding boxes | Classes | Locations |
|--------------|-------------|--------|----------------|---------|-----------|
| EPIC KITCHENS (Damen et al., 2022) | 11,500,000 frames, 454,158 bounding boxes, 290 classes, 32 kitchens, 4 cities | 11,500,000 | 454,158 | 290 | 32 kitchens in 4 cities |
| DAVIS (Deng et al., 2009) | Object segmentation dataset | - | - | - | - |
| CDnet2014 (Deng, 2009) | Moving object detection dataset | - | - | - | - |
| VOT (Everingham et al., 2015) | Object tracking dataset | - | - | - | - |
| MOT (Oreifej et al., 2012) | Object tracking dataset | - | - | - | - |
| Sports-1M (Kuznetsova et al., 2020a) | Segment-level annotation dataset for sports actions | - | - | - | - |
| HMDB-51 (Rybak and Dudczyk, 2020) | Segment-level annotation dataset for human action categories | - | - | - | - |
| TRECVID (Lin et al., 2014) | Video retrieval and indexing dataset | - | - | - | - |
| Caltech Pedestrian Detection (Deng et al., 2009) | Pedestrian detection dataset | - | - | - | - |
| PASCAL VOC (Real et al., 2017; Russakovsky et al., 2015) | Object detection dataset | - | - | - | - |
| Semi-supervised/ unsupervised methods (Chen et al., 2022) | Works based on semi-supervised or unsupervised methods | - | - | - | - |

### 3.2.2 Epic Kitchens dataset

A dataset known as EPIC KITCHENS was made available in 2018 (Real et al., 2017). It includes 11,500,000 frames with 454,158 bounding boxes covering 290 classes from 32 distinct kitchens spread across four cities. However, general video object detection is limited in its kitchen scenario. In addition, the following other datasets are available that represent particular applications: the object segmentation dataset DAVIS (Varghese et al., 2020), the moving object detection dataset CDnet2014 (Wang et al., 2014), the object tracking datasets VOT (Kristan et al., 2015) and MOT (Dendorfer et al., 2021), the segment-level annotation Sports-1M dataset (Karpathy et al., 2014), the segment-level annotation HMDB-51 dataset (Kuehne et al., 2011) for different human action categories, the video retrieval and indexing dataset TRECVID (Awad et al., 2017), the pedestrian detection dataset Caltech Pedestrian Detection (Dollár et al., 2009; Hamza et al., 2020), and the object detection PASCAL VOC dataset (Everingham et al., 2010, 2015). Furthermore, certain studies that employ unsupervised or semi-supervised approaches have been examined in references.

**Table 6**    CDnet dataset description

| Feature | Description |
|---|---|
| Dataset name | 2014 CDnet |
| Video source | Realistic, camera-captured (without CGI) videos |
| Recording devices | Low-resolution IP cameras, higher resolution consumer-grade camcorders, commercial PTZ cameras, near-infrared cameras |
| Spatial resolutions | Ranges from 320 × 240 to 720 × 486 |
| Lighting conditions | Diverse lighting conditions |
| Compression parameters | Vary, leading to varying levels of noise and compression artefacts |
| Video duration | 900 to 7,000 frames |
| Radial distortion (for low-res IP cameras) | Noticeable radial distortion |
| Hue bias | Different cameras have different hue biases due to varying white balancing algorithms |
| Exposure adjustment | Some cameras apply automatic exposure adjustment, resulting in global brightness fluctuations over time |
| Frame rate | Varies between videos, often due to limited bandwidth |
| Change detection methods | The dataset does not favour a certain family of change detection methods over others due to the wide range of settings under which the videos were captured |

### 3.2.3 Extended CDnet 2014 dataset

Similar to the 2012 CDnet, the 2014 CDnet contains a variety of authentic, camera-captured (without CGI) indoor and outdoor videos. A variety of cameras were utilised to capture these recordings, including consumer-grade camcorders with higher resolutions, commercial PTZ cameras, low-resolution IP cameras, and near-infrared cameras. Consequently, the 2014 CDnet features videos with spatial resolutions ranging from 320 × 240 to 720 × 486. The amount of noise and compression flaws in different videos is very different because of the different lighting conditions and compression settings that were used. The video durations range between 900 and 7,000 frames. Videos captured by IP cameras with a low-resolution exhibit discernible radial distortion. The hue bias of various cameras varies as a result of the white balancing algorithms that are implemented. Certain cameras utilise automatic exposure adjustment, which causes time-dependent fluctuations in global luminance. Frame rate also differs between videos, frequently due to bandwidth constraints. Due to the diverse array of conditions in which these videos were captured, the extended 2014 CDnet dataset does not exhibit a bias towards a specific category of change detection techniques (Wang et al., 2014; Zeng et al., 2018).

### 3.2.4  DOTA

DOTA is a big dataset for finding objects in pictures taken from above. It can be used to test and improve object detectors that use overhead pictures. The pictures come from a number of different sensors and systems. The images range in size from 800 × 800 to 20,000 × 20,000 pixels and show items with a lot of different sizes, shapes, and orientations. Professionals who know how to read aerial pictures label the instances in DOTA images with a random eight-degree quadrilateral (Xia et al., 2018). We will keep adding to DOTA so that it gets bigger and better as things change in the real world. There are now three kinds of it:

- The first version of DOTA has 15 main groups, 2,806 images and 188,282 instances. In DOTA-v1.0, the training set, validation set, and testing set make up 1/2, 1/6 and 1/3 of the whole.

- DOTA-v1.5 employs identical images to DOTA-v1.0, with the addition of annotations for exceedingly minute instances (less than 10 pixels). Additionally, the classification 'container crane' is introduced. In total, it comprises 403,318 instances. Identical to DOTA-v1.0 in terms of both picture count and dataset partitioning. Released in conjunction with IEEE CVPR 2019, this version was made available for the 2019 DOAI Challenge on Object Detection in Aerial Images.

- DOTA-v2.0 adds new aerial, GF-2 Satellite, and Google Earth photographs to its collection. With 11,268 photos and 1,793,658 instances, DOTA-v2.0 features 18 common categories.

The new categories 'airport' and 'helipad' are added, making it even better than DOTA-v1.5. There are training, validation, test-dev, and test-challenge sets made up of the 11,268 pictures in DOTA. The percentage of the training and validation sets is lower than the test set to prevent the overfitting issue. Additionally, test-dev and test-challenge are our two test sets. There are 268,627 instances and 1,830 images in the training set. There are 81,048 instances and 593 photos in the validation. For training and validation sets, we made the image and ground truth sets available. There are 353,346 instances and 2,792 images in test-dev. The ground truths were not disclosed; just the pictures were. 1,090,637 instances and 6,053 photos make up test-challenge. The ground truths and images of the test-challenge will only be accessible during the challenge itself.

## 4    Methods for images-based object detection

Computer vision has a powerful application called object detection. It involves identifying and categorising objects within an image. Object detection methods can be classified into two categories – traditional and deep learning-based detectors. Figure 1 depicts this classification. Deep learning-based detectors can be further categorised into parts like 'two-stage detector' and 'one-stage detector'.

**Figure 1** Different models chart for different types of object detection (see online version for colours)

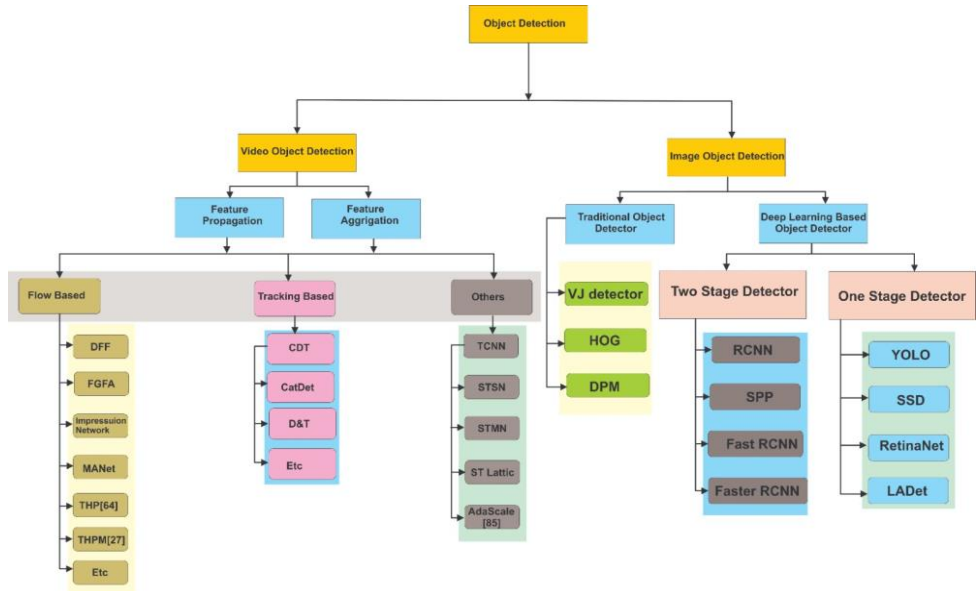

### 4.1 Traditional detectors

In the past, object recognition methods used features that were made by hand. A lot of researchers only used complex features vectors back then because more advanced ways to describe images were not fully developed yet, and computers were not very fast. Objects are found by traditional detectors using methods such as the Viola-Jones (VJ) detector, HOG, SIFT, and others. The old ways of finding objects worked because the features that described them were carefully chosen to find the important parts of the picture. Using feature representation and a classifier, the astounding outcome was obtained on the Pascal VOC dataset (Kiran et al., 2022).

### 4.1.1 Viola-Jones detectors

In order to identify a human face, VJ detector was the first classical object detection method to do so 21 years ago. Unrestricted use of this method for face detection has been demonstrated (Wu et al., 2020). VJ employs a sliding window or other simple detection method. This method makes it easy to detect a human face in a window by checking and scaling all of the image's pixel positions. This algorithm has successfully identified faces at that point in time without any limitations (Setta et al., 2022). Since 'Paul Viola and Micheal Jones' made substantial contributions to the field of face detection, the approach is named after the authors. Although the method itself is simple, the calculations required to complete it were beyond the capabilities of the computers available at the time. Consequently, while the detection speed is quick, the training speed is slow for this technique. There are now three distinct phases to this whole procedure:

- *Integral image:* this method has been utilised to enhance the speed of box filtering, albeit it is computational in nature. In order to figure out the rectangular features, the integral representation of the input picture is used as a starting point for the next steps. Haar wavelet has been used by the VJ detector to show picture features.

- *Feature selection:* integration was used to pick out the small and important features. A search was done using the AdaBoost method to find a small group of good features.

- *Detection cascade:* this final stage of the method removes an object's backdrop from an image using cascade classifiers. An item is subjected to more computations than the entire image. As a result, it concentrates exclusively on the object (faces) and lowers the computational overhead in the background.

### 4.1.2 Histograms of oriented gradients

The histograms of oriented gradients (HOG) descriptors, which have enhancements in the scale-invariant feature transform (SIFT) and shape context, were proposed in 2005 by Dalal and Triggs. It uses patches to construct the histogram of edges; a patch can comprise any kind of item, person, or significant background. Pre-processing, such as translating to a predetermined aspect ratio of 1:2 and resizing the patch to that size if it exceeds $64 \times 128$ pixels, is done on certain patches using this technique. Preprocessing involves the use of gamma correction. With this method, the gradient histogram – that is, the vertical and horizontal gradients – is computed. The Sobal operator can be used for this, and the following formula can then be used to determine the gradients' magnitude and direction. Three colour channels are assessed if there are colors in the supplied image. As a result, gradients are computed for each pixel in the image for three colour channels. The largest gradient from the three-channel and matching angle is then chosen for additional processing. To get the final result, a gradients vector histogram is made and normalised to the vectors. Subsequently, data is sent to machine learning techniques, such support vector machines (SVM), which are used to train classifiers. To eliminate the most unnecessary bounding box of an item, non-maximum suppression (NMS) is used. To identify several objects of a certain class, HOG is employed. By repeatedly rescaling the image while maintaining the window size constant, it is able to detect all distinct object sizes. The pedestrian detection feature of HOG has seen extensive use.

### 4.1.3 Deformable part-based model

Felzenszwalb et al. (2008) presented deformable part-based model (DPM), the pinnacle of conventional object detection techniques. When it comes to HOG detectors, DPM is the upgraded version. In order to detect threats, DPM uses a divide-and-conquer approach. In this context, divide means training and conquer means inference. Training involves learning in a way that facilitates object decomposition. In order to calculate the overall inference, the results of detecting the object's pieces are integrated. The root filter, the part filter, and the spatial model are the three parts that make up this model. A root filter effectively encompasses the entire item inside its detection window. Consequently, the 'region feature vector' is filtered using a set of predetermined weights. Part filter: to hide the fine details in an image, you can use a multiple-part filter. The locations of the part filters are scored in relation to the root using a spatial model.

Furthermore, Felzenszwalb et al. (2008) expanded on this work by creating the 'star model'. After some time, Girshick et al. (2015) transforms the 'star model' into a mixture model to identify real-world objects under various but notable modifications. Weakly supervised learning has been used in the mixed model to configure the part filters. Put another way, weakly supervised learning has been used to automatically learn component filters. P. Felzenszwalb and R. Girshick received a 'lifetime achievement' award from PASCAL VOC in 2010.

## 4.2   Deep learning-based object detectors

Object detection has emerged as a prominent domain within the field of computer vision due to technological advancements. Prior to 2010, the traditional technique was used for object detection. Conventional detectors have several drawbacks, such as producing an enormous number of redundant proposals among their many created. Numerous false positives were the outcome of these repetitive suggestions (Wu et al., 2020). Thus, the CNN architecture's proposal brought about a significant shift in the fields of deep neural networks and object identification. In order to enhance object detection accuracy and minimise errors, a deep neural network has been utilised to abstractly represent the image's features. Due to the massive quantity of data required for training, deep neural architecture is notoriously time-consuming. Object detection makes use of a variety of backbone architectures, including AlexNet, VGG Net, Google Net, and others (Zhang and Hong, 2019). There are two types of deep learning-based object detection algorithms: one with two stages and another with just one. Objects in images can be detected using two-stage detectors, which frequently produce state-of-the-art results or excellent accuracy on existing datasets. On the other hand, compared to one-stage detectors, these ones have slower inference speeds. Compared to two-stage detectors, one-stage detectors produce the desired outcome much more rapidly, making them ideal for use in real-time object identification.

### 4.2.1   One-stage detector

In order to get the final result, a one-stage detector does not conduct any intermediate tasks. Consequently, it sets the standard for easier and speedier architecture. When a bounding box needs to be assigned a specified position, one-stage detectors like YOLO, RetinaNet, SSD, etc., come into play. Consequently, detectors learn based on the positions of specific objects.

#### 4.2.1.1   You only look once

The first stage detector, you only look once (YOLO), was presented by J. Redmon, S. Divvala, R. Girshick and colleagues in 2015 (Redmon et al., 2016). YOLO employs many techniques for identification and validation. Figure 2 depicts the YOLO architecture. This method is quick since it makes use of a fixed grid detector. This method uses a single neural network to recognise items across the entire image. The entire image is segmented into fixed sections, and the probability and bounding box of each object are computed from those regions. In YOLO, the class probabilities for several bounding boxes were predicted using a single CNN. Here, entire photographs are used for training. YOLO makes object predictions using S × S grid-based convolutional layers.

Object detection and boundary prediction are handled by these distinct grids on the input image.

You may find YOLO in a number of different iterations, including YOLOv2, tiny-YOLO-voc1 [forest fire detection (Yuan and Lu, 2018)], YOLOv3 (Zhao et al., 2016), YOLOv4, and YOLOv5. YOLOv's accuracy drops to rock bottom when the input picture size differs from the training image size and it struggles to recognise tiny objects (Redmon et al., 2016). Hence, YOLOv2 is debuted with an architecture based on Darknet-19, which has 19 convolution layers, five max pool layers, and one softmax layer (Redmon and Farhadi, 2018). By modifying the scaling activation function, batch normalisation decreased overfitting. The YOLOv2 image resolution increased from $224 * 224$ to $448 * 448$. Additionally, the accuracy of detecting tiny objects was improved with a $13 \times 13$ feature map. Additionally, the YOLOv2 version's performance was enhanced using the single shot multibox detector (SSD) and region proposal network. YOLOv3, which is built on the Darknet-53 architecture, has better accuracy for smaller objects than its predecessors (Redmon and Farhadi, 2017).

**Figure 2**  Architecture of YOLO detection system (see online version for colours)

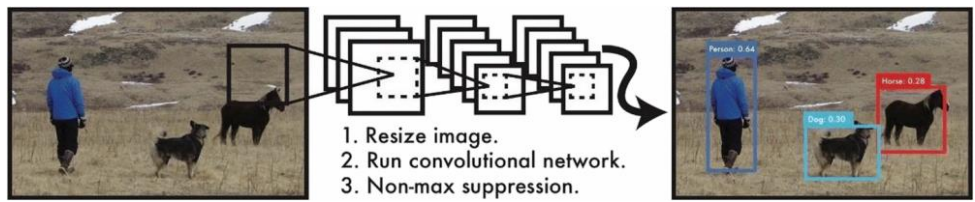

*Source:*  Fu et al. (2017)

YOLOv3, which offers the advantages of a multi-label classifier in object detection, uses a logistic classifier in place of the SoftMax function. The YOLOv3's primary innovation is its ability to detect objects at three distinct scales. Feature pyramid network (FPN) in YOLOv3 enables the network to learn objects with varying sizes. Traffic object detection from video is done using the EYOLOv3 (Sudha and Priyadarshini, 2020). The most recent version of YOLOv4, which has been enhanced by 10% in mAP, has been shown to be the quickest object detection due to advancements in YOLO (Bochkovskiy et al., 2020). The latest iteration of YOLOv4, which has improved the mAP by 10%, has been shown to be the quickest object detection due to advancements in YOLO. Yolov4 divides the current layer into two separate sections using a CSPDarknet53 feature extraction architecture. There are two methods: one that involves running each input through the convolution layer and the other that does not. Ultimately, the outcomes from both sections are combined. By isolating the most important context feature in YOLOv4, spatial pyramid pooling improves accuracy. Transforming the Darknet framework into the PyTorch framework is what YOLOv5 is all about as of late (Kuznetsova et al., 2020b; Nepal and Eslamiat, 2022). Darknet provides extensive management of the network's activities and is mostly written in C. However, PyTorch allows for more configuration-free control of low-level operations and is developed in Python. The main characteristics of YOLOv5 are data augmentation, auto-learning bounding box anchors, lightweight models, and quick performance. Four distinct model sizes are included in the YOLOv5 release:

1    YOLOv5s (small)

2    YOLOv5m (medium)

3    YOLOv5l (large)

4    YOLOv5x (exceptionally huge).

### 4.2.1.2 *Single shot multibox detector*

Liu et al. (2016) presented a novel method called SSD. It is the second one-stage detector technique in the era of deep learning. SSD employed multi-reference and multi-scale representation to improve the accuracy of small item recognition. The SSD only operates on the top layers, whereas the previous object identification method operated on five separate layers. This is the sole distinction between the two methods. High detection accuracy real-time fire detection is another application for SSD (Wu and Zhang, 2018). The SSD head and the backbone model comprise the two components of the SSD architecture. The backbone model, a feature map extractor and pre-trained classification network, is the initial component. The first portion, which consists of stacking several convolutional layers, is applied first, and then the second part. The output is shown in this section as the bounding box (BB) over the object that was detected. These convolutional layers detect different items in the image. It is a quick and effective object identification model that can identify objects in several categories. Eventually, tiny SSD was introduced, and on the VOC07 dataset, it proved to be more dependable than tiny YOLO (Womg et al., 2018). Using optimised SSD, the authors can employ SSD to optimise the algorithm for detecting things like wheels and cars (Fu et al., 2020). ResNet-101 (backbone) used the deconvolutional single shot detector (DSSD), which is an expanded version of the faster RCNN.

Additionally, this method includes two modules: the deconvolution and prediction modules (Fu et al., 2017; Mehmood et al., 2024b). The prediction module incorporates the residual block into each prediction layer and then adds the output of both the prediction layer and the residual block element by element. The deconvolution model improved feature strength by increasing feature map resolution. In the deconvolution layer, many types of objects of varying sizes are predicted. Using the ILSVRC CLS-LOC dataset, the model was trained using the ResNet-101 backbone architecture. The purpose of the experiment is to demonstrate the efficacy of the suggested model on the MS COCO and PASCAL VOC datasets. An improvement of 2.2% is achieved on the VOC2007 dataset using the deconvolution and prediction module. Scientists have been working on ways to enhance current detectors and increase their accuracy in real-time settings. To address issues brought on by small objects, scale variation, and complicated backdrops, Lu et al. (2021) introduced the AF-SSD (Attention and feature fusion SSD). The multiscale receptive field (MRF) CNN module, which displays the extent of the region where the feature map pixels are captured on the original picture, was utilised by this structure. When the receptive field size is big, the feature map contains more global information, and when it is small, the feature map contains more specific information. In order to capture features at several scales, this module has been specifically built to widen the receptive field.

### *4.2.1.3 RetinaNet*

A new loss function has been proposed by Lin et al. (2017) to replace the current conventional cross-entropy loss in order to address the issue of class imbalance during the model's training. The accuracy of a one-stage detector is slower than that of two-stage detectors. As a result, RetineNet has presented a quick and accurate single-stage object detector that works with both small- and dense-scale objects. With two subnetworks and a backbone design, it functions as a single, unified network for object detection at various scales. Input images of any size can be processed by the backbone architecture, which then calculates the convolutional feature map. A second subnetwork carries out bounding box regression, while the first subnetwork classifies objects based on the backbone architecture result. As illustrated in Figure 9, a one-stage detector's performance is enhanced by means of two sub-networks.

Top-down and bottom-up paths, as well as a classification and regression subnetwork, are the four primary parts of RetinaNet (Figure 3). The first technique is known as the bottom-up approach, and it involves using ResNet as the foundational architecture to calculate feature maps at various screen sizes. In the second section, lateral linkages and top-down paths are implemented using feature pyramid network (FPN). From images of any size with a single resolution, this aids in building a multi-scale and rich feature pyramid. The classification subnetwork, which is RetinaNet's third component, then makes probability predictions for each detected item class and anchor box. The final step is to determine the BB's offset in relation to the anchor boxes. Similar to two-stage detectors, the focal loss (FL) function preserves accuracy while having a high detection speed. When it comes to stability, RetinaNet really shines with larger objects (Nguyen et al., 2020). FL enhances the cross-entropy loss that was suggested as a solution to the foreground-background class imbalance issue in the one stage detector paradigm. At each pyramidal layer of the suggested network, there are thousands of anchor boxes.

**Figure 3**    Architecture of one stage detector RetinaNet with FPN (see online version for colours)

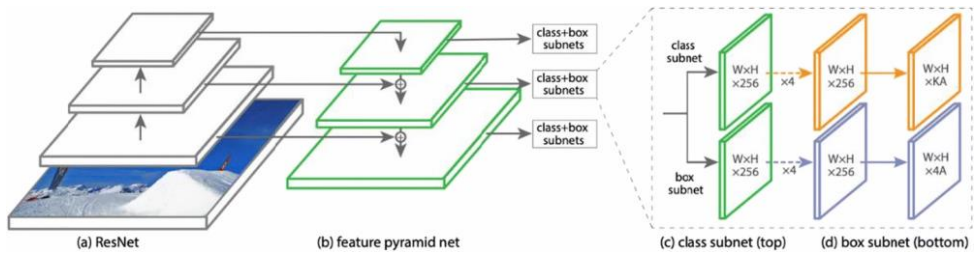

*Source:*   Everingham et al. (2010)

It finds extensive use in object detection in aerial and satellite photos. R4, short for 'refined single-stage detector with feature recursion and refinement for rotating objects', is a new end-to-end detector created by Sun et al. (2020). It employs RetinaNet as its basis network to recognise items from dense distribution, bigger aspect ratio, and category imbalance.

### 4.2.1.4   Lightweight and adaptive network for multi-scale object detection

In 2019, Zhou et al. (2019a) came up with the idea of LADet, a lightweight and adaptable network for multi-scale object recognition that can deal with the problems that come up when there are differences in scale. Two modules, the AFPM and the light-weight classification function module (LCFM), have been used by LADet to accomplish object detection (Figure 4 shows the architecture of LADet). The suggested architecture extracts the multi-level feature from the input image by using the DenseNet-169 as its foundation. Additional retrieved characteristics were passed into SFPM, where the feature maps produced corresponding semantic information. The feature fusion module (FFM) and the adaptive channel-wise feature refinement (ACFR) are the two subcomponents of this module that are illustrated in Figure 4. Scale normalisation ensures that all feature maps of pyramid levels in the FFM are of the same resolution. To further fuse the feature maps, numerous outputs of scale normalisation are concatenated.

**Figure 4**   Architecture of LADet detector RetinaNet with AFPN (see online version for colours)

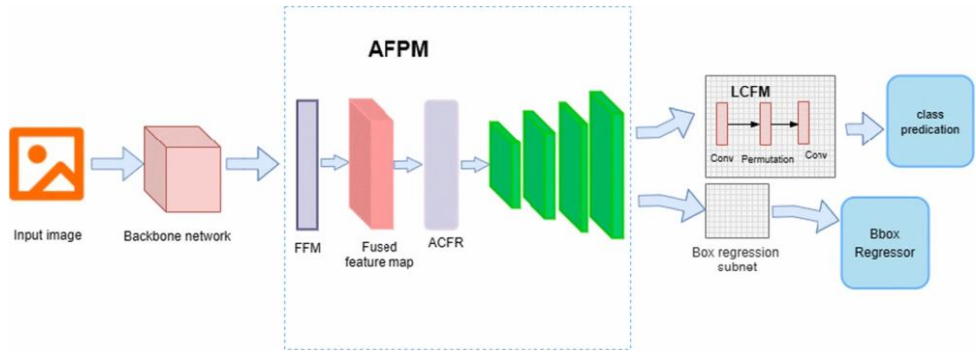

After that, the ACFR model adds to the FFM output by creating supplementary data. Using the learnt feature maps, dense bounding boxes and classification scores are produced. With the use of structure-sparse kernels, the AFPM has employed two convolution layers. In order to estimate the 'high-rank kernel' or the 'original dense', this module additionally used the permutation operation between these layers. The detected object and its class are likely to be present, according to the predictions made by the classification module. This technique has improved accuracy and speed compared to previous techniques. It includes using AFPM to get multi-level FM from the backbone network and making pyramid FM with semantic information that is paired. Then, the classification subnet (LCFM) and regression subnet (box) are utilised.

### 4.2.2   Two-stage detectors

There are two steps to this process: first, detecting the objects, which involves proposal creation; and second, making predictions about the objects based on these proposals. The initial step is to use detectors to identify regions of all the objects. At the same time that the image detector creates regions with a high recall rate, objects are considered to be part of at least one of these regions. Deep learning models are used in these detectors' second stage to carry out classification. The regions that are formed either have a background or objects with predetermined labels. In spite of this, the model suggested in the first step may also help to improve localisations. This section also covers the widely used two-stage detector approaches.

### 4.2.2.1  *Region-based convolutional neural network (RCNN)*

With CNN, Girshick et al. (2015) developed RCNN, a two-stage method for object detection. In order to detect objects, the RCNN method goes through three stages: extracting the region, computing CNN features, and last, classifying the region. During the region extraction process, a total of 2,000 regions were created utilising the selective search method, which involved cropping and wrapping. Since objects exist on all sizes, the selective search method quickly locates them all using computer power. We classify comparable areas according to their size, texture, colour, and form. To calculate CNN, first choose areas. Then, once each region is downsized to a fixed size, it is submitted to CNN to extract features. For the purpose of region classification, support vector machines (SVMs) employ non-maximum suppression (NMS) on each class to eliminate areas with intersection over union (IOU) values greater than the learnt threshold.

Consequently, in order to identify the object in the picture, conventional detectors relied on feature descriptors that were hand-crafted. In this comparison, the image hierarchical characteristics were produced using a deep neural network, and object detection was accomplished by capturing information from multiple layers of varying scales. Figure 4 shows the processes carried out by the R-CNN architecture, as mentioned in the preceding paragraph (Hu and Liu, 2024). The object is both classified and its bounding box is displayed using this method. Another issue with RCNN is that it uses a deep convolution network to extract features from each proposal independently, which results in a lot of unnecessary calculations. Because of this, the training and testing of RCNN required a significant amount of time. Moreover, the three independent steps of this technique made it challenging to find the globally optimal solution. Lastly, selective search underperforms in complex image backgrounds when generating proposals because it only uses low-level signs.

### 4.2.2.2  *Fast region-based convolutional neural network*

Girshic (2015) presented a new architecture called Fast RCNN and further improved RCNN and SPPNet. It uses the convolution layer to generate feature maps from input images, which are taken as a whole.

It gets rid of the multi-level sharing layer and only uses a single-layer grid. The 'region of interest (ROI)' pooling layer is used to get a feature vector with a set length. This feature vector is then used with the 'spatial pyramid layer (SPP)' and its single pyramid layer. Before the output layer, a series of 'fully convolution (FC) layers' work on each feature vector. The final FC output is split into two layers that are similar to each other: softmax and bounding box. Probabilities for each object class and one background object class have been generated via the softmax layer as shown in Figure 5. Four actual values have been produced by the Bbox layer to create the box surrounding the anticipated object. With this method, the classifier and Bbox regression are not trained separately. Python and C++ (using Caffe) are used to build Fast RCNN (Girshick, 2015).

**Figure 5** Network structure of SPPNet (see online version for colours)

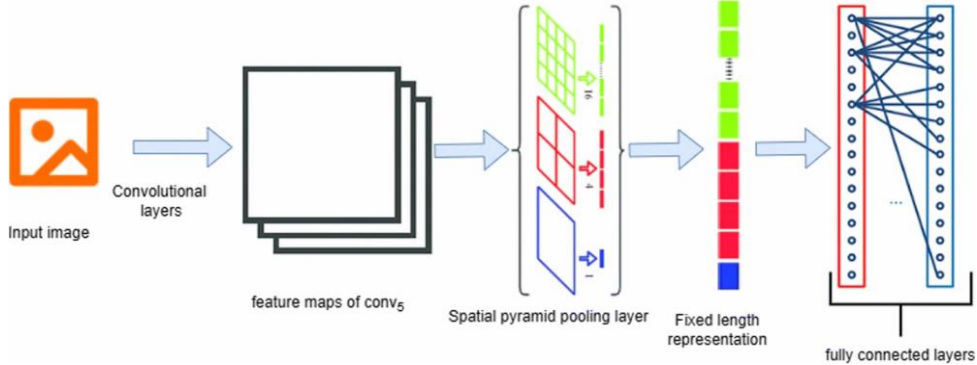

It took the best parts of both RCNN and SPP-Net and put them together. However, it is a little slow because it has to identify proposals. But it saves money on extra storage room, makes things more accurate, and works better.

### 4.2.2.3 Faster region-based convolutional network method

The possible boxes are made in a number of different ways, such as using selective search, edge boxes, and so on. Nevertheless, the object detection approach's efficiency is not being enhanced by these strategies. For this reason, a new technique called RPN in Faster RCNN was introduced in 2015 by Ren et al. As seen in Figure 6, RPN generates the bounding box and bounding box score for each object on the feature map using the sliding window method.

An anchor box is a bounding box that has been constructed and shares a common aspect ratio. The proposals are moved to the complete convolution layer, which includes the softmax and regression layers, after being reduced to a fixed size. Applying ReLu over the output window increases the nonlinearity in the (n × n) Convolution window. We were able to train an object detection algorithm using this new end-to-end architecture. Authors identify the items by modifying the design of these methods (Han et al., 2019).

The primary goal of Mask RCNN, a two-stage process based on a deep neural network, is to resolve the instance segmentation issue (Kaur and Singh, 2022). It is an expansion of the Faster RCNN detectors, with the addition of a mask network branch module for segment prediction within region of interests (ROIs). This technique can be used for both segmentation of instances and object detection in parallel. Using this method, different items are separated from images and videos. To put it simply, it receives an image as input and outputs bounding boxes, masks, and classes for the items that are discovered. The initial level of this network takes an input image and produces region suggestions where the object may have existed. Stage 2 involves item class prediction, bounding box refinement, and pixel-level mask creation. The mask RCNN architecture is linked to the FPN backbone in two steps.

**Figure 6**   Architecture of Fast RCNN (see online version for colours)

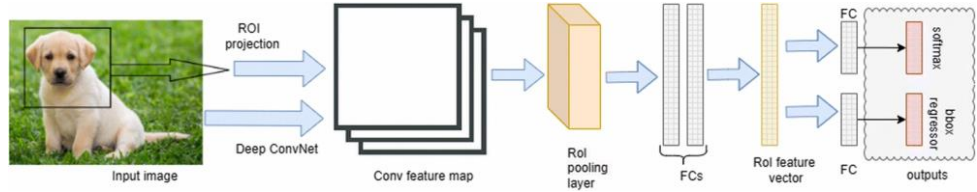

### 4.2.2.4   *Spatial pyramid pooling networks (SPPN)*

In 2015, He et al. read and took into consideration the theory of spatial pyramid matching (SPM) to create a novel network structure called SPP-Net in order to address the shortcomings of RCNN. This constraint is artificial because CNN is applied to a fixed size input image in RCNN. Certain objects are deformed or cropped due to fixed size limitations. To increase accuracy, SPP Net has eliminated their requirement. The new network architecture makes advantage of the SPP layers, which generate the prescribed length vector representation that the fully linked convolution layer accepts. Unlike RCNN, this detector uses the entire image only once to extract the features. The entire image is used to construct a feature map using the deep convolutional network. Moreover, fixed-length feature vectors were extracted by the SPP layer using the feature map. As seen in Figure 5, feature maps are split into three fixed-size grids, or M × M by SPP layer, for the numerous values of M. Feature vector generated from every grid cell. The feature vectors are then combined and fed into the bounding box regressor and SVM classifiers. This method is 20 times faster than RCNN, but it maintains the same accuracy. Similar to RCNN, it has two shortcomings: first, it ignores all layers prior to the completely convolution layer; second, SPPNet requires multistage training. This approach works well without causing unwanted distortion or information loss because it operates on multiple scales and aspect ratios.

## 5   Methods for video-based object detection

For video object detection, different ways of capturing the link between time and space are being looked into so that the video's properties can be used to their fullest. Some papers (Guo and Gao, 2012; Yadav and Singh, 2016; Oreifej et al., 2012) looked at the old ways of doing things. These works are mostly designed by hand, which makes them less accurate and less resistant to noise sources. Recently, deep learning techniques have attempted to address these drawbacks. Figure 7 shows the characteristics that differentiate various video object detectors, including those that use flow-based, LSTM-based, attention-based, tracking-based, and other approaches that use temporal information and aggregate features extracted from video snippets (Chen et al., 2020c; Zhu et al., 2017b; Wang et al., 2018; Zhang and Kim, 2019).

- *Flow-based approaches:* these methods utilise optical flow to propagate feature maps or motion cues from key frames to non-key frames, thereby reducing redundant computations and improving accuracy under motion blur or defocus. Examples include deep feature flow (DFF) and flow-guided feature aggregation (FGFA).

- *Tracking-based approaches:* these methods combine detection and object tracking, using trackers to predict object locations in subsequent frames and detectors to correct or refine predictions. This integration leverages temporal continuity to improve efficiency and robustness, particularly under occlusion or rapid object motion.

- *Attention-based approaches:* these rely on attention mechanisms to selectively emphasise important spatial regions or temporal frames, filtering out less informative content. By focusing computational resources on the most relevant features, they achieve improved precision in crowded or complex environments.

- *Recurrent/LSTM-based approaches:* these methods explicitly model temporal dynamics using recurrent neural networks (e.g., LSTM or GRU). They capture long-term dependencies across frame sequences, making them effective in handling gradual motion changes or extended temporal contexts.

- *Hybrid or other approaches:* these combine two or more of the above strategies (e.g., flow + tracking, or attention + LSTM) or introduce alternative mechanisms such as deformable convolutions or adaptive scaling. They are often designed to strike a balance between accuracy and computational efficiency in real-world surveillance applications.

A more in-depth explanation of these methods is as follows:

## 5.1 Flow-based

The flow-based methods make use of optical flow in two distinct ways. According to (Zhu et al., 2017b), using deep feature flow for video recognition (DFF) – an acronym for the term – is the first way to reduce computation. Similarly (you can find more acronyms in the sources you mentioned), optical flow is used to send information from key frames to non-key frames. In the second way, optical flow is used to make each frame better by using the time-spatial information between frames that are close to it, as explained in (Zhu et al., 2017a) (FGFA). Despite having slower speeds, the second approach has been shown to have superior detection accuracy. Consequently, efforts were made in Hetang (2023) (impression network), Zhu et al. (2018) (THP) and Fujitake and Sugimoto (2022) (THPM) to incorporate both of these approaches using the same methodology. Using the pixel-level temporal-spatial information, an optical flow method was presented in Blachut and Kryjak (2022) to determine the difference between succeeding frames. In order to estimate optical flow, FlowNet's deep learning model was used in Dosovitskiy et al. (2015).

### 5.1.1 DFF

Modern object detection algorithms developed for still images are not well suited to identifying objects in moving images for the reasons already given. Consequently, the following constraints were considered when developing the DFF method:

1   the amount of time needed to compute feature map extraction for each video frame

2   the propagation of feature maps between frames

3    the degree to which features extracted from two neighbouring frames are similar.

In order to obtain the feature map from sparse key frames, Zhu et al. (2017b) employed ResNet-101, a convolutional neural subnetwork. Instead of feature extraction from the feature map on key frames as ResNet-101 did, non-key frame features were acquired through the process of warping the feature map on key frames using the flow field produced by FlowNet (Dosovitskiy et al., 2015). Figure 3 shows the structure that was drawn. This method speeds up the process of finding objects on non-key frames. DFF was able to obtain a mAP accuracy of 73.1% on the ImageNet VID dataset (Russakovsky et al., 2015) when it was operating at 20 frames per second. This is in comparison to the baseline accuracy of 73.9% when operating at 4 frames per second. There was a significant improvement made to the method's practical application to the visual object detection.

**Figure 7**    DFF (deep feature flow) framework (see online version for colours)

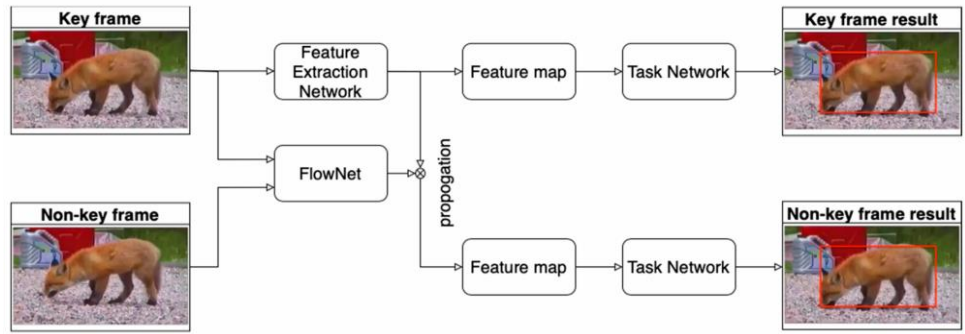

*Source:*    Qiang et al. (2020)

### 5.1.2  FGFA

There was an idea in Zhu et al. (2017a) to use flow-guided feature aggregation (FGFA) to make recognition more accurate when there is motion blur, rare poses, video defocus, and other issues. ResNet-101 (He et al., 2016) was used to pull out feature maps from each frame of the movie. To improve the feature maps of a current frame, the optical flow network's motion information was used to warp the feature maps of close frames to the current frame.

A small sub-network was used to obtain a new embedding feature by inputting the warped feature maps and the extracted feature maps on the current frame.

In order to calculate the weights, a similarity measure based on the cosine similarity metric was used in conjunction with this new embedding feature. Subsequently, the weights were used to aggregate the features. In essence, the sub-network committed to shallow detection utilised the combined feature maps to ascertain the ultimate detection outcome for the present frame. In Figure 4, the FGFA framework is displayed. In comparison to DFF, FGFA obtained an accuracy of 76.3% mAP at 1.36 fps using the ImageNet VID dataset.

**Figure 8** Flow-guided feature aggregation (FGFA) framework (see online version for colours)

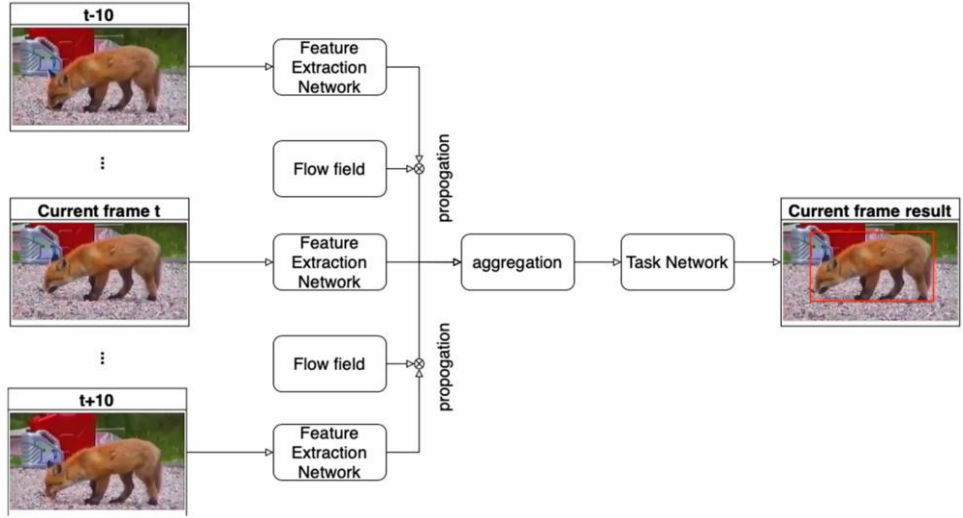

*Source:* LeCun et al. (2015)

### 5.1.3 Impression network

The FGFA feature fusion method made recognition more accurate, but it took a lot longer to do the calculations. However, feature propagation approaches sacrificed detection accuracy for increased computing efficiency. 2017 saw the development of a network that was given the name 'impression network' (Hetang, 2023), with the purpose of simultaneously enhancing performance in terms of both accuracy and computing speed. The idea that individuals recall previous frames even when they see a fresh one led to the combination of sparse key-frame features with extra key frames in an effort to improve detection accuracy. By acquiring feature maps of non-key frames by a frame, the accuracy of detection was also enhanced. Additionally, feature maps of non-key frames were obtained by employing a technique similar to the one described in reference (Guo and Gao, 2012) and a flow field. By feature propagating the characteristics of the non-key frames, the efficiency of the inference computation was increased. In contrast to the technique described in Zhu et al. (2017a), the feature aggregation method for the key frames obtained the weight maps for each localisation using a small fully convolutional network. The impression network got 75.5% mAP precision on the ImageNet VID dataset when it was running at 20 frames per second. The weight maps for each localisation, which was different from how it was explained in Zhang and Kim (2019). The impression network got 75.5% mAP precision on the ImageNet VID dataset when it was running at 20 frames per second.

### 5.1.4 THP

Another combination method, THP, was presented in addition to impression network in Zhu et al. (2017b). While the previous approaches all made use of key frames with fixed intervals, this one added temporally adjustable key frame scheduling to further improve the speed-accuracy trade-off. Key frames with a fixed interval present a challenge when it comes to maintaining quality control over the frames. By employing temporally adaptive key frame scheduling, the percentage of sites exhibiting poor optical flow quality was utilised to dynamically modify the fixed interval key frames. A value exceeding a specified threshold T would indicate that the current frame has undergone a significant quantity of change relative to the key frame preceding it. The feature maps were subsequently extracted from the present frame, which was subsequently chosen as the subsequent key frame.

The mAP achieved an accuracy of 78.6% and operated at 13.0 and 8.6 frames per second on the NVIDIA Titan X and K40 GPUs, respectively, according to the findings in Guo and Gao (2012)(]. )t faster velocities, the mAP decreased marginally to 77.8% when the T was altered (22.9 fps on Titan X and 15.2 fps on K40). Titan X outperformed the feature propagation (He et al., 2016) and aggregation (Zhang and Kim, 2019) based winning entry in the 2017 ImageNet VID challenge with a 76.8% mAP at 15.4 frames per second. In addition, Guo and Gao (2012) demonstrated superior performance in terms of accuracy and speed.

### 5.1.5  THPM

In the same way, THPM (Yang et al., 2019) offered a simple network structure for finding objects in videos. On key frames, a light picture object detector is used. The cutting-edge, portable. The network that serves as the backbone is MobileNet (Howard et al., 2017). Light flow networks take feature maps from key frames and send them to non-key frames so they can be detected. The features can be efficiently aggregated between key frames with the help of a flow-guided gated recurrent unit (GRU) module. On mobile devices (such as the Huawei Mate 8 from HUAWEI TECHNOLOGIES CO., LTD., China), THPM attains 60.2% mAP at a frame rate of 25.6 fps using the ImageNet VID dataset.

### 5.2  Tracking-based

A lot of different methods have been created to find things at set interval frames and follow them between frames, because tracking is thought to be a good way to use temporal information (Yang et al., 2019; Howard et al., 2017). The new methods in He et al. (2016) and Luo et al. (2019) can find interval frames on the fly and A lot of different methods have been created to find things at set interval frames and follow them between frames, because tracking is thought to be a good way to use temporal information (Yang et al., 2019; Howard et al., 2017). The updated approaches in Yang et al. (2019) and Luo et al. (2019) adaptively identify interval frames and follow the remaining frames.

### 5.2.1  CDT

In Kim and Kim (2016), a system called CDT was shown that combines detection and tracking to find objects in videos. This framework included an object detector, a forward tracker, and a backwards tracker. The picture object detector initially spotted things. The forward tracker tracked each detected object, while the backwards tracker stored unseen objects. Throughout the process, the object detector and the tracker worked together to keep track of when things showed up and when they disappeared. In Mao and Kong (2019), a system called CDT was shown that combines detection and tracking to find objects in videos. There was an object detector, a forward tracker, and a backwards tracker in this system. At first, the image's object detector was able to identify things. The forward tracker continued to follow each item it had spotted, while the backwards tracker recorded any objects that had not been discovered. During the entire procedure, the tracker and object detector worked together to handle objects' arrival and departure.

### 5.2.2 CaTDet

In Mao and Kong (2019), another competingly efficient framework called CaTDet was introduced. Figure 9 depicts this setup, which consists of a detector and a tracker. With a tracker, CaTDet can confidently anticipate where items will be in the following frame. The steps in processing CaTDet are:

1   each frame is sent to a proposal network, which returns possible ideas in the another framework called CaTDet that uses very little computing power was shown in Feichtenhofer et al. (2017)

2   the tracker makes a very good guess about where an object will be in the next frame

3   the outcomes of the tracker and the proposal network are mixed and sent to a refinement network to get the measured object information.

**Figure 9**   CaTDet framework (see online version for colours)

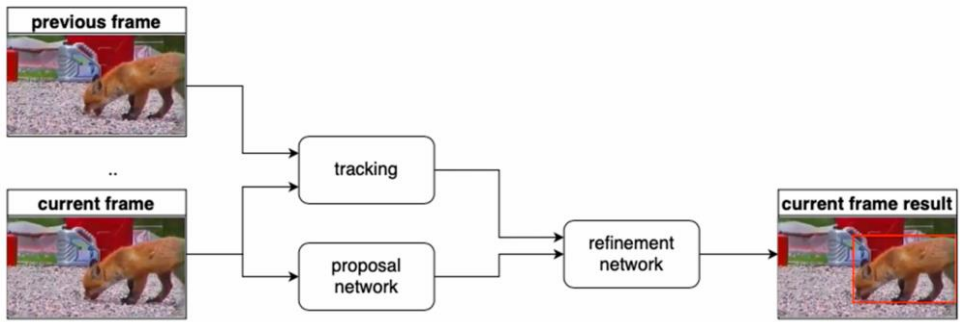

*Source:*   Mao and Kong (2019)

To be more precise, a tracker was employed to forecast the locations on the subsequent frame using the previous data, based on the assumption that items spotted in one video frame would presumably emerge in a subsequent frame. If new items showed up in a current frame, proposals were found using a proposal network similar to RPN that was fast on computers. A tracker also used the time information to guess where it would be in the future to deal with problems like motion blur and occlusion. Following the combination of the tracker and proposal networks, a refinement network was applied to the resulting data. The refinement network limited its refinement to the regions of interest in order to conserve computation time without compromising accuracy.

Contemporary methodologies for object detection and tracking in video, related to CDT and CaTDet, incorporate intricate multistage components. A framework employing a ConvNet architecture was deployed in a straightforward yet impactful manner in Dai et al. (2016) by concurrently carrying out tracking and detection. In particular, the initial R-FCN (Ren et al., 2015) was utilised to derive the shared feature maps for tracking and detection. Then, RPN based on anchors was used to gather proposals in each frame (Luo et al., 2019). The ultimate detection was achieved by pooling RoIs (Luo et al., 2019). To be more specific, the architecture was extended by introducing a regressor. The regression maps from both frames, which were sensitive to location, were combined with correlation maps to serve as the input for a RoI tracking module. The module then produced the box relationship that existed between the two frames. In order to determine the effectiveness of the system described in Feichtenhofer et al. (2017) for video object detection, it was tested on the ImageNet VID dataset and achieved an accuracy of 82.0% mAP.

### 5.2.3  *D&T*

In the same way, discussed a framework (D or T) that utilises a scheduler network to ascertain the action (detecting or tracking) performed on a specific frame (refer to Figure 10). The idea for this system came from the discovery that tracking objects works better than finding them. The basic frame skipping method, which involves finding on fixed-interval frames and following on intermediate frames, did worse on the ImageNet VID dataset than the light, simple-structured scheduling network. Furthermore, crucial frames were selected using the adaptive approach in (TRACKING ASSISTED). The identification of critical frames required the implementation of a precise detection network, while the monitoring module provided assistance for the detection of non-critical frames.

**Figure 10**    D or T framework (see online version for colours)

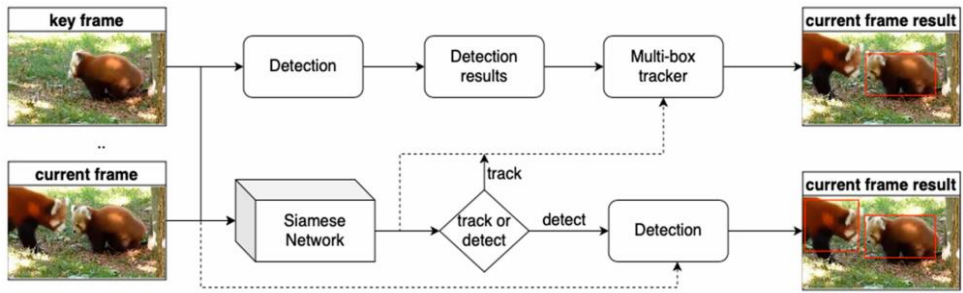

*Source:*  Luo et al. (2019)

### 5.3  *Other methods*

Besides the models mentioned above, some methods are shown that are a mix of several of the methods mentioned above (Kang et al., 2017; Chen et al., 2019). The techniques described in Kang et al. (2017) are based on tracking and optical flow. Both the attentional LSTM method (Chen et al., 2019) and the TSSD method (Chen et al., 2018) are based on LSTM techniques approach using LSTM and focus.

### 5.3.1 STSN

The literature also includes some other strategies (Bertasius et al., 2018; Du et al., 2017; Zhu et al., 2020b; Jiang et al., 2019). Both Bertasius et al. (2018) and Du et al. (2017) cover techniques for improving and aligning feature maps. However, in order to find the optimal compromise between speed and accuracy, the approach presented in Ahmad et al. (2025) investigated the impact of the input image size. A technique described in Bertasius et al. (2018) is the spatial-temporal sampling network (STSN). By using this method, feature maps can be aligned between neighbouring frames. Similar to the FGFA technique described in Chin et al. (2019), this method is based on the premise that detecting objects in a single frame is challenging due to the presence of noise sources such motion blur and defocused video. Multiple frames are employed to augment features in order to gain improved performance. Deformable convolution, in contrast to FGFA, employs the defocus technique for aligning feature maps. Multiple frames are used to add features in order to gain improved performance. In contrast to FGFA, which use the optical flow technique for aligning feature maps, feature alignment in Bertasius et al. (2018) is achieved by the use of deformable convolution. Initially, a network for extracting sharing features is utilised to extract feature maps from both the current frame and the neighbouring frames. Next, the two feature maps are merged together based on their channels, and then a deformable convolution operation is applied. Aligning the feature maps requires a second deformable convolution operation, and the output of the first one serves as the output set. Also, similar to FGFA, feature aggregation yields enriched feature maps. Unlike FGFA, STSN uses flexible convolution to automatically line up the features of two frames next to each other as shown in Figure 11. Results are good, but it is not as user-friendly as the optical flow approach. Even without using the optical flow data, the experimental results showed that STSN still outperformed FGFA in terms of mAP (78.9% vs. 78.8%). Another interesting finding is that STSN outperformed the D&T baseline (Zhu et al., 2017a), with a performance of 78.9%.

**Figure 11**  STSN (spatiotemporal sampling networks) framework (see online version for colours)

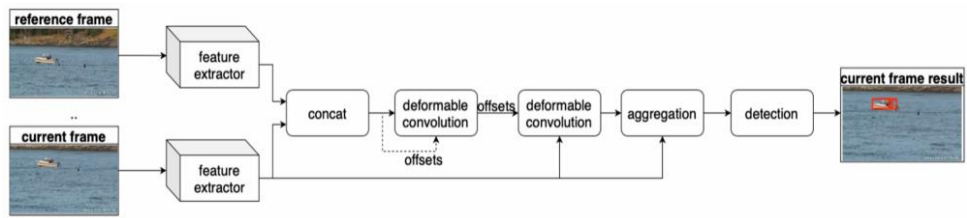

*Source:*  Bertasius et al. (2018)

### 5.3.2 STMN

In contrast to Bertasius et al. (2018), the spatial-temporal memory network (STMN) was looked at in Du et al. (2017). It used deformable convolution to send time information and an RNN design with a STMM to include long-term time information. For object detection in video, the end-to-end STMN model's long-term data and aligns motion dynamics.

STMM serves as the fundamental component of STMN, a convolutional recurrent computation unit that utilises to the fullest extent the weights that were pretrained using static image datasets like ImageNet (Feichtenhofer et al., 2017). It is hard to learn from video datasets because they do not have a lot of different objects in the same group. This design is needed to solve that problem. At time step t, STMM gets the feature maps of the current frame and the information from all the frames that came before it in the spatial-temporal memory MIT 1. Subsequently, the spatial-temporal memory MIT for the present time step is revised. The memory M, utilised for both classification and bounding box regression, is created via bidirectional feature aggregation using two STMMs. This allows for simultaneous recording of information from both recent and past frames. As a result, data from many video frames is used to propagate and aggregate the feature maps. After being tested on the ImageNet VID dataset, STMN has reached the state-of-the-art accuracy level.

### 5.3.3   AdaScale

The propagation and aggregation of feature maps is the starting point for all of the algorithms outlined above. Video object detection was investigated from a different angle in Zhu et al. (2020b). Zhu et al. (2020b) examined the impact of input image size on video object detection performance, which is comparable to that in Zhu et al. (2020a). Additionally, it was discovered that down-sampling photos might occasionally achieve superior accuracy. In this light, the AdaScale framework for adaptively choosing the input image size was suggested. In order to determine the optimal size or scale for the next frame, AdaScale uses the data from the current frame. For one reason, things are getting better: there are fewer false positives. Also, this is done to increase the number of true positives by cutting down the items that are too big for the detector.

In Zhu et al. (2020b), the best scale (the pixels on the image's shortest side) is found using a set of fixed scales S (in this case, S = [600, 480, 360, 240]). Furthermore, a loss function, which incorporates both the classification loss and the regression loss, is employed as an evaluation statistic for comparing results across several scales. The background regression loss will probably be zero. Therefore, when comparing results at different sizes using the loss function directly, the image scale with less foreground bounding boxes is acceptable. To get around this issue, we employ a new metric called the loss function to compare photos of varying sizes.

It examines a consistent number of foregrounds bounding boxes selected at varying sizes. It is the smallest number (m) on all scales that determines the number of containing boxes needed to figure out the loss. Based on a ranking of the expected loss, we select the first m bounding boxes on each scale to get the optimal foreground bounding boxes. The optimal scale is the one that results in the lowest loss, denoted as m. The assumption is that the size information is contained in the deep features' channels, as was made by R-FCN (Dai et al., 2016) while working on deep features for bounding boxes regression. The best way to determine the scale to use is to construct a scale regressor that makes use of deep features. When tested on the ImageNet VID and mini-YouTube-BB datasets, AdaScale outperformed single-scale training by 1.6 times and testing by 1.8 times, respectively, leading to mAP gains of 1.3% and 2.7%. Also, when DFF (Zhu et al., 2017b) was added, the speed went up by 25% while keeping the same mAP on the ImageNet VID dataset.

## 6 Result analysis

The ImageNet VID dataset (Russakovsky et al., 2015) is used to test the performance of most video recognition methods. The comparison is shown in Tables 4, which are related to post-processing. LSTM methods are used to add the sequence information. These methods were influenced by LSTM-based techniques in natural language processing. Flow&LSTM (Zhang and Kim, 2019) attained the maximum accuracy of 75.5% among the LSTM models. Looking Quickly and Slowly (Fujitake and Sugimoto, 2022) produced results with rapidity but poor precision. LSTM implements long-term information in a straightforward fashion. Because the input and forget gates' sigmoid activation is rarely fully saturated, there is delayed state decay and a lack of long-term dependency. This means it is hard to keep the whole previous state during the update. It has been shown that attention-based methods can also effectively find objects in videos. With ResNeXt-101 serving as the backbone, MEGA (Chen et al., 2023) attained the maximum accuracy of 84.1% mAP in the attention-related group. It was able to reach a very high level of precision while moving at a speed that was relatively quick. The characteristics that are contained inside the proposals that are generated are compiled by attention-based approaches. It takes less time to compute as a result of this. To a certain extent, the performance is dependent on the effect of RPN. This is due to the fact that the proposals only use the features that are contained inside them. The utilisation of information that is more complete is fairly challenging in this context.

In the group that is based on tracking, the approaches are aided by tracking. Zhu et al.'s (2017a) D&T loss was able to obtain 75.8%mAP. The utilisation of temporal information through the use of a detector that is helped by a tracker is an effective strategy utilised in tracking. On the other hand, it is not capable of directly resolving the issues that are caused by motion blur and video defocus. In light of the fact that the performance of the detector is dependent on the success of the tracking, the detector component is prone to tracking mistakes. Furthermore, there are additional approaches that can be used independently, such as TCNN (Kang et al., 2017), STSN (Bertasius et al., 2018) and STMN (Du et al., 2017).

### 6.1 Result comparison for video-based object detection

Table 7 provides an overview of different types of frameworks used in object detection, each with its specific backbone and their corresponding mean average precision (mAP%) performance metric. The discussed types include flow-based, LSTM-based, attention-based, tracking-based, and others. Among the flow-based models, the flow-guided feature aggregation (FGFA) and looking fast and slow frameworks utilise ResNet-101 and Inception-ResNet backbones, showcasing mAP% scores of 78.4 and 80.1, respectively. The Interleaved + Quantization + Async approach achieves a slightly lower mAP% of 59.3. In the LSTM-based category, variations of the MobilenetV2-SSDLite + LSTM framework demonstrate diverse mAP% scores, with the highest being 64.1 for MobilenetV2-SSDLite + LSTM (=1.4). The attention-based models, such as PSLA, SELSA, and RDN, utilising ResNet-101, ResNeXt-101, and ResNeXt-101 + DCN backbones, achieve competitive mAP% scores ranging from 78.6 to 85.4. Tracking-based approaches, like D&T (=10) and D&T (=1), leverage ResNet-101 and Inception V4 backbones, attaining mAP% scores of 79.8 and 82.0, respectively. Finally, the other category includes models like STSN and STMN, which employ ResNet-101 + DCN and ResNet-101 backbones, achieving mAP% scores of 80.4 and 80.5, respectively.

**Table 7**    Video-based object detection state-of-the-art comparison

| Type | Framework | Author | Backbone | mAP% |
|---|---|---|---|---|
| Flow-based | FGFA | LeCun et al. (2015) | ResNet-101 | 78.4 |
| | Looking Fast and Slow | Wei and Kehtarnavaz (2018) | Inception-ResNet | 80.1 |
| | | | Interleaved + Quantisation + Async | 59.3 |
| LSTM-based | MobilenetV2-SSDLite + LSTM (=1.4) | Zhao et al. (2016) | MobilenetV2-SSDLite | 59.1 |
| | MobilenetV2-SSDLite + LSTM (=1.0) | | | 50.3 |
| | MobilenetV2-SSDLite + LSTM (=0.5) | | | 45.1 |
| | MobilenetV2-SSDLite + LSTM (=0.35) | Redmon et al. (2016) | ResNet-101 | 80.8 |
| | OGEMN | Tao and Busso (2018) | ResNet-101 + DCN | 81.6 |
| Attention-based | PSLA | Brena et al. (2020) | ResNet-101 | 80.54 |
| | | | ResNeXt-101 | 84.50 |
| | RDN | Wei et al. (2020) | ResNet-101 | 78.60 |
| | | | ResNeXt-101 | 84.70 |
| | SELSA | Tao and Busso (2020) | ResNet-101 | 83.80 |
| | | | ResNeXt-101 | 85.40 |
| Tracking-based | D&T (=10) | Yuan and Lu (2018) | ResNet-101 | 79.8 |
| | D&T (=1) | | Inception V4 | 82.0 |
| Others | STSN | Varma and Sreeraj (2013) | ResNet-101 + DCN | 80.40 |
| | STMN | Bhangale et al. (2020) | ResNet-101 | 80.50 |

Choosing the 'better' type depends on specific use cases and requirements. Flow-based models seem promising with high mAP% scores, but the choice between looking fast and slow and FGFA might depend on the specific demands of the application. LSTM-based models demonstrate adaptability with different MobilenetV2-SSDLite + LSTM configurations, offering a trade-off between accuracy and efficiency. Attention-based models consistently achieve high mAP% scores, with SELSA and RDN standing out. Tracking-based approaches exhibit competitive results, especially D&T (= 1) with a mAP% of 82.0. Ultimately, type of object detection framework depends on the application's unique needs. If real-time processing is critical, a lightweight model with acceptable accuracy may be preferred. For applications where high precision is essential, attention-based models with strong mAP% scores may be the best choice. It is essential to strike a balance between speed, accuracy, and resource constraints, ensuring that the selected framework aligns with the practical demands and limitations of the specific use case. Additionally, continuous evaluation and testing in the target environment are crucial to validate the chosen framework's performance under real-world conditions.

The results reported in Table 7 are compiled from original benchmark studies; the corresponding source references are indicated for each framework for transparency.

## 6.2 Results comparison for image-based object detection

Table 8 provides a comprehensive comparison of various object detection models, evaluating their mean AP (mAP%) on MS-COCO and Pascal-VOC 2007 datasets, along with their frames per second (FPS) performance. Each model exhibits unique characteristics, contributing to the diversity in their mAP scores and efficiency in real-time processing.

**Table 8**     Summary of different object detection (Pascal Titan X GPU) performances on MS-COCO and Pascal-VOC07

| Models | Authors | mAP% (MS-COCO) | mAP% (Pascal-Voc 2007) | FPS |
|---|---|---|---|---|
| RCNN | Chen et al. (2020b) | – | 66 | 0.1 |
| SPPNet | Redmon and Farhadi (2018) | – | 63.10 | 1.0 |
| Fast RCNN | | 35.90 | 70.00 | 0.5 |
| Faster RCNN | Zhou et al. (2019b) | 36.20 | 73.20 | 6.0 |
| Mask RCNN | Zhu et al. (2017a) | – | 78.20 | 5.0 |
| YOLO | He et al. (2015) | – | 63.40 | 45 |
| SPSLASD | Dai et al. (2016) | 31.20 | 76.80 | 8.0 |
| YOLOv2 | Shrivastava et al. (2016) | 21.60 | 78.60 | 67 |
| YOLOv3 | Wei and Kehtarnavaz (2019) | 33.00 | – | 35.0 |
| SqueezeDet | Chen et al. (2020c) | – | – | 57.2 |
| SqueezeDet+ | | | | 32.1 |
| CornerNet | Zhu et al. (2017b) | 69.2 | – | 4.0 |

Notably, the Faster RCNN family, including Faster RCNN and Mask RCNN, achieves competitive mAP scores on both datasets, showcasing advancements in region proposal networks and instance segmentation. YOLOv2 and YOLOv4 excel in real-time processing, demonstrating high FPS values while maintaining respectable accuracy. Efficient single-shot detectors like SSD strike a balance between accuracy and speed, making them versatile for various applications.

Models like SqueezeDet and SqueezeDet+ prioritise lightweight architectures, delivering competitive FPS for real-time applications. CornerNet, with its corner-based approach, achieves high mAP on MS-COCO, emphasising accuracy in complex scenes. The original RCNN and SPPNet, while pioneering, exhibit relatively lower FPS, indicating potential challenges in real-time deployment.

Faster RCNN, YOLOv2, YOLOv4, and SSD offer versatile solutions, with trade-offs between accuracy and real-time processing capabilities. Lightweight models like SqueezeDet and SqueezeDet+ cater to resource-efficient applications. The selection should align with the desired balance between mAP, FPS, and the practical needs of the intended use case, such as precision object detection or real-time responsiveness.

The performance statistics in Table 8 are collected from published works; the corresponding source references are indicated for each framework for transparency.

Table 8 provides a comprehensive overview of various object detection models, categorised into two-stage detectors and one-stage detectors, along with specialised models. The evaluation metrics include average precision (AP) across different difficulty levels (easy, medium, hard) and a combined metric (AP(E), AP(M), AP(H)). The analysis reveals several trends and insights into the performance of these models.

### 6.2.1 Comparison of video object detection methods

1   *Two-stage detectors:*

- Two-stage detectors, such as Faster RCNN with FPN and Mask RCNN, tend to achieve higher average precision (AP) values across difficulty levels.

- The incorporation of feature pyramid network (FPN) proves effective in handling multi-scale features, enhancing performance in both easy and challenging scenarios.

2   *One-stage detectors:*

- One-stage detectors, including YOLOv2, YOLOv3, and various SSD models, showcase a trade-off between speed and accuracy.

- These models typically achieve competitive AP values, making them suitable for real-time applications where efficiency is crucial.

3   *Specialised models:*

a   *SqueezeDet and SqueezeDet+:*

- Lightweight models like SqueezeDet and its enhanced version, SqueezeDet+, demonstrate impressive AP values, emphasising the effectiveness of streamlined architectures in achieving high accuracy.

b   *CornerNet models:*

- CornerNet models, utilising a corner-based approach, show competitive AP values, indicating the potential of alternative detection strategies.

4   *Overall considerations:*

a   *Backbone influence:*

- The choice of backbone models, such as VGG16, ResNet-101, DarkNet53, and SqueezeNet, plays a crucial role in shaping the performance of the respective detectors.

b   *Performance trade-offs:*

- Different models cater to varied requirements, with some prioritising accuracy (e.g., Mask RCNN), while others emphasise speed and efficiency (e.g., YOLO series).

c   *Specialised approaches:*

- SqueezeDet and CornerNet models present alternative methodologies, showcasing that innovation in model architecture can lead to notable improvements in object detection performance.

Two-stage detectors are often preferred for tasks requiring high precision, while one-stage detectors and specialised models cater to scenarios demanding real-time processing. The continuous evolution of model architectures and techniques highlights the dynamic nature of the object detection field, offering a range of solutions tailored to diverse applications.

### 6.3 Discussion and future prospective

Although deep learning applications have made significant progress in object detection and classification, some outstanding issues still need to be addressed in the future. These include improving the accuracy of these systems and addressing the limitations in their capabilities.

Several techniques have been developed to identify and categorise small vehicles in datasets that are publicly available. To improve the accuracy of both classification and localisation of small vehicle objects, despite occlusions, variations within and between classes, differences in lighting, surroundings, and other factors, it is necessary to modify the model architecture in the following ways:

Several studies have investigated multi-task joint optimisation because of the link between vehicle object categorisation and detection tasks. This involves the utilisation of a combination of information from many models. Re-identification of individuals (Ahmad et al., 2025), grouping and recognising human actions (Mehmood et al., 2021), detecting dangerous objects (Mehmood et al., 2024c), detecting objects quickly (Mehmood et al., 2025), recognising and tracking vehicles with multiple tasks (Mujtaba et al., 2024), and estimating the pose of vehicles with multiple tasks (Alawaji et al., 2024). Furthermore, a range of methodologies have been integrated to optimise the efficacy of the designs.

- *Scale and size alteration:* small objects exhibit more noticeable variations in scale and size, necessitating the use of multi-scale object classifiers and detectors to increase robustness to changes. Advanced algorithms like ResNet, Inception, MobileNet, and AlexNet can be used for detecting and categorising objects of varying sizes. FPN creates feature maps on different scales and different GAN-based representations for small and big objects with less work for the detectors and classifiers to do on the computer. The network shows how to make a useful feature stack for finding objects in different situations. To identify items in a variety of ways, you need to combine cascade architecture and scale distribution estimation.

- *Spatial correlations and contextual modelling:* object detection and image classification rely heavily on the spatial distribution of objects. In order to identify probable object locations, region proposal generation and grid regression techniques are used. On the other hand, position-sensitive score maps in RFCN do not regulate global structure information, and these methods do not consider the rectification between several proposals and object classes. Collaborative approaches combining methods like sequential reasoning problems and subset selection are used to tackle these issues.

- *Cascade architecture:* a cascade network is made up of a series of detectors that are constructed in stages. The inability to optimise distinct ConvNets is a limitation of existing cascade designs caused by the fact that training fixes earlier steps in cascades. Therefore, optimising the ConvNet cascade design from beginning to end is of the utmost importance.

- *Weakly supervised and unsupervised learning:* labelling a lot of bounding boxes by hand can be time-consuming and wasteful. To solve this issue, various architectures can be put together to work very well by using image-level supervision to give object classes that match object boundaries and regions. This method makes detection more flexible and cuts down on work costs.

- *Model optimisation:* when working with deep learning applications and schemes, it is important to optimise the models in order to balance accuracy, speed, and memory usage. This can be done by selecting an optimal detector and classifier. To do so, it is important to keep in mind the needs of your audience and use simple, everyday language. Keep sentences short and direct, using the active voice whenever possible. This will help ensure that the most important information is presented clearly and logically.

- *Detection or classification in videos:* one of the most important challenges for autonomous driving and video surveillance is the real-time object recognition and classification in videos. Traditional object classifiers or detectors are typically created for detecting and classifying images only, without considering the correlations between video frames. Therefore, it is imperative to focus on enhancing detection or classification performance by exploring spatial and temporal correlations. This is an important area of research that needs to be further developed to address the current limitations of object detection and classification in videos.

- *Lightweight classification or detection:* despite the development of lightweight architectures, they continue to be troubled by categorisation errors in models. Therefore, there is a need to improve the detection accuracy. While significant progress has been made in recent years, there is still a need to balance the speed of detection and classification accuracy.

## 7  Conclusions

This paper is a detailed review that includes major developments, achievements, and limitations associated with the use of deep learning (DL) approaches in object detection from video and image sources. These studies recorded that the different deep-learning approaches are very successful in giving improved detection accuracies, and processing speeds, as well as being able to model adaptability to the complexities in surveillance environments. The detailed analyses presented on the preprocessing techniques and the feature engineering underline the instrumental role that benchmark datasets play in making a collaborative effort toward fine-tuning model performances. In spite of good progress, the object detection domain brought forth quite demanding problems: those to deal with occlusions, motion blur, and yet without mechanisms to exploit temporal information for dynamic object tracking. These issues further raise the requirement for pursuing new research directions in designing efficient, robust, and scalable deep learning solutions able to perform real-time analysis with precise detection capability. We also report the challenges that have been motivating technical developments taking place in the last few years. This paper therefore suggests possible future directions in the object detection field. It guides the development of neural networks and related learning frameworks by recommending useful insights and directions for their further development.

It is also important to note that the present work does not introduce a new detection framework. Instead, it aims to systematically review, compare, and highlight the strengths and weaknesses of existing techniques. Acknowledging this limitation ensures clarity of scope and emphasises that the novelty of this review lies in its comprehensive synthesis and identification of open challenges, which can guide future research efforts toward developing innovative detection architectures.